\documentclass[letterpaper, 10 pt, journal, twoside]{ieeetran}  

\IEEEoverridecommandlockouts                              

\usepackage{multicol}
\usepackage[bookmarks=true]{hyperref}
\usepackage{caption}
\usepackage{subcaption}
\usepackage{amsfonts}
\usepackage{multirow}
\usepackage{algorithm}
\usepackage{algorithmic}
\usepackage{amsmath}
\usepackage[normalem]{ulem}
\usepackage{graphicx}
\usepackage{amssymb}
\usepackage{wrapfig}

\usepackage{soul}
\usepackage[dvipsnames]{xcolor}

\begin{document}

\title{CE-MRS: Contrastive Explanations for Multi-Robot Systems}

\author{Ethan Schneider, Daniel Wu, Devleena Das, Sonia Chernova
\thanks{This work was supported by Army Research Laboratory under Grants W911NF-17-2-0181 (DCIST CRA).}
\thanks{The authors are with the Institute for Robotics and Intelligent Machines, Georgia Institute of Technology, Atlanta, GA, USA {\tt\footnotesize \{eschneider32, chernova\}@gatech.edu}}
}%


\maketitle

\begin{abstract}
As the complexity of multi-robot systems grows to incorporate a greater number of robots, more complex tasks, and longer time horizons, the solutions to such problems often become too complex to be fully intelligible to human users. In this work, we introduce an approach for generating natural language explanations that justify the validity of the system's solution to the user, or else aid the user in correcting any errors that led to a suboptimal system solution.  Toward this goal, we first contribute a generalizable formalism of contrastive explanations for multi-robot systems, and then introduce a holistic approach to generating contrastive explanations for multi-robot scenarios that selectively incorporates data from multi-robot task allocation, scheduling, and motion-planning to explain system behavior. Through user studies with human operators we demonstrate that our integrated contrastive explanation approach leads to significant improvements in user ability to identify and solve system errors, leading to significant improvements in overall multi-robot team performance.
\end{abstract}
\begin{IEEEkeywords}
    Design and Human Factors, Human Factors and Human-in-the-Loop, and Multi-Robot Systems
\end{IEEEkeywords}

\section{INTRODUCTION}
\vspace{-.1cm}
\IEEEPARstart{H}{eterogeneous} multi-robot systems offer the possibility to solve complex problems in a variety of industries and environments {\cite{antonyshyn2023multiple}}.  To provide a solution, such systems must solve three subproblems, determining which robot should perform which task (\textit{task allocation}), in what order (\textit{scheduling}), and how (\textit{motion planning)}.  As the complexity of multi-robot systems grows to incorporate a greater number of robots, more complex tasks, and longer time horizons, the solutions to such problems often become too complex to be fully intelligible to human users.  Despite this, human operators are routinely tasked to review, validate, and provide final approval for complex task allocation and scheduling solutions prior to robot deployment \cite{Adams100Robots, adams2023human}. The phenomenon of increasingly complex black-box models is not unique to multi-robot systems and is commonly found across Machine Learning and Artificial Intelligence research, leading to the emergence of Interpretable Machine Learning (IML) \cite{gilpin2018survey} and Explainable AI (XAI) \cite{fox2017explainable} subfields which seek to provide a human-interpretable explanation for the decision making of complex models.

Research in social sciences has shown that humans seeking an explanation for complex phenomena typically utilize a \textit{contrastive} approach \cite{miller2019explanation,lipton2017inference}.  Instead of seeking a summary of a solution, or identifying a complete causal chain of reasoning that led to some outcome, humans instead probe the solution by asking a question framed as a contrast against some other expected outcome.  Such a query may be phrased as \textit{``Why is robot1 performing task t instead of robot2?''} or \textit{``Why is action a in the solution?''} (implying a comparison to a solution without the given action).  Human studies show that when providing explanations in such cases to each other, humans provide partial explanations instead of full ones, focusing an explanation on the key factors that caused the given output instead of another \cite{ruben2015explaining}.
 
We are interested in providing explanations that aid operators in interpreting solutions generated by complex multi-robot systems that incorporate task allocation, scheduling, and motion planning into its decision making. Prior XAI work has addressed this challenge by introducing techniques for generating explanations for task allocation \cite{zehtabi2023contrastive,zahedi2023why}, scheduling \cite{ludwig2018scheduling,cyras2019scheduling}, and motion planning \cite{gyevnar2024causal} independently.  However, recent work in the multi-robot community has shown that the close \textit{interdependency} between these three subproblems (i.e., determining which robots should perform which tasks affect the timing/schedule of those tasks, and in turn, the motion plans required for their execution) is most effectively addressed by holistic solutions that consider all three challenges together \cite{neville2021interleaved,messing2022grstaps,neville2023ditags}. We argue that explanations, similarly, must have the ability to incorporate information across subproblems in order to most accurately represent the system's decision making to the operator.

In this work, we present Contrastive Explanations for Multi-Robot Systems (CE-MRS), a holistic explanation framework for multi-robot systems. Specifically CE-MRS allows users to ask a hypothetical query about the system solution to construct a foil solution. CE-MRS then provides a contrastive natural language explanation specifying:
\begin{itemize}
    \item whether the user's foil solution is feasible; if not, why.
    \item how the model's solution compares to the user's foil solution with respect to various performance metrics (e.g. time or resources).
\end{itemize}
The end-user can then repeat this process to receive multiple explanations for different foils.
\textbf{Through this approach, our objective is to leverage natural language explanations to justify the validity of the system's solution to the user, or else aid the user in correcting any errors that led to a sub-optimal system solution.}  

Our work makes the following contributions. First, we formalize contrastive explanations for multi-robot systems. Second, we contribute a holistic approach to generating contrastive explanations for multi-robot scenarios through CE-MRS, which utilizes select information from the multi-robot task allocation, scheduling, and motion-planning sub-problems. Finally, we validate our approach in a 22-participant in-person user study in which each participant evaluated the quality and validity of 6 multi-robot solutions in a search-and-rescue domain. Our results show that by utilizing information from all three sub-problems, we can better explain the reasoning for multi-robot solutions. 





\section{Related Works}
\vspace{-.1cm}
In this section, we situate our work in its relation to prior research in XAI and multi-agent planning (MAP).

\subsection{Multi-Agent Planning}
The problems of task allocation, scheduling, and motion planning have been studied extensively in the context of multi-robot systems \cite{gerkey2004mrta,korsah2013comprehensive,antonyshyn2023multiple}, often under the umbrella of multi-agent planning \cite{torreno2017survey}. Our work addresses a variant of this problem as it relates to single-task (ST) robots, multi-robot (MR) tasks, and time-extended allocation (TA), as defined in Korsah et. al \cite{korsah2013comprehensive}. 
In this work, we contribute a generalizable approach for providing explanations for integrated multi-robot task allocation, scheduling, and motion planning systems.  Our approach is domain-independent and generalizes to multiple underlying representations; in our experiments we adopt the representation and problem formulation of Neville et al. \cite{neville2021interleaved}, as presented in Section \ref{sec:prob_des}.

\subsection{Explainable AI Planning}
A branch of XAI, Explainable AI Planning (XAIP) provides explanations of sequential decision making models \cite{fox2017explainable}. 
Prior work in XAIP has primarily focused on providing explanations in single-agent domains, which fall into one of three categories \cite{chakraborti2020emerging}: algorithm-based explanations \cite{greydanus2017visualizing, guo2023advising, koul2018learning}, model reconciliation \cite{chakraborti2017soliloquy}, and plan-based explanations \cite{rosenthal2016verbalization, chakraborti2019planning}. \textit{Algorithm-based explanations} provide explanations for sequential decision making models, e.g. single-agent deep reinforcement learning (RL) \cite{greydanus2017visualizing}, multi-agent RL \cite{guo2023advising}, or actor critic techniques \cite{koul2018learning}. \textit{Model reconciliation} based explanation techniques seek to reconcile detected differences between the mental models of the robot and user based on user queries \cite{chakraborti2017soliloquy}. Examples of \textit{plan-based explanations} include using a verbalization space to narrate a segment of a motion plan \cite{rosenthal2016verbalization} or constructing a minimal set of causal links to justify actions in a plan \cite{chakraborti2019planning}.

Recently, there has been a shift in additionally exploring explanations for multi-agent domains, proposed as Explainable decisions in Multi-Agent Environments (xMASE) \cite{Kraus_2020}. Recent work in multi-agent model reconciliation propose using natural language and AR visualizations to summarize plans to end-users \cite{luebbers2024explainable}. Multi-agent plan-based explanations include: summarizing multi-robot plans by visualizing disjoint segments of the plan \cite{kottinger2022conflict} and using plan argument schemes to justify a multi-agent plan \cite{mahesar2023argument}. Recent work from {\cite{brandao2022taxonomy}} has provided a taxonomy for what questions and explanations users prefer in the multi-agent path finding domain. Within that taxonomy, our work addresses plan-centered questions and problem-based metric explanations.

\subsection{Explainable Task Allocation}
For multi-robot systems, a common challenge is to allocate a set of robots to a set of tasks by taking into consideration the robot's traits, task requirements, and relative locations of the robots and tasks \cite{neville2021interleaved,messing2022grstaps,neville2023ditags}. A recent work in xMASE introduces CMAoE \cite{zehtabi2023contrastive}, a domain independent approach for providing tabular constrastive explanations by utilizing features of an objective-function for multi-agent systems. 
Another recent work introduces AITA \cite{zahedi2023why}, a negotiation-aware explicable task allocation framework which provides graphical contrastive explanations by providing minimal information from the human-agent preferences to refute the end-user's foil.  
Notably, unlike prior work \cite{zehtabi2023contrastive,zahedi2023why}, we do not assume the system's plan represents the optimal solution to the problem. Instead, we consider the solution as optimal \textit{given the user-provided problem specification}, and allow for the possibility that the problem specification may be impacted by \textit{human error}. 

\section{Contrastive Explanations for MAP}
\vspace{-.1cm}
Contrastive explanations provide an answer to the question \textit{``Why P and not Q?''}, where \textit{P} represents the algorithmic solution, and \textit{Q} represents the user's alternative suggestion or \textit{foil} \cite{miller2021contrastive}.  Examples of user foils include questions such as \textit{``Why is action a in the solution?''} and \textit{``Why is robot1 performing task t instead of robot2?''}. 
Several works within XAIP have demonstrated the use of contrastive explanations for model reconciliation, helping users correct their mental models and understand the correctness of solutions within the context of single-agent scenarios \cite{chakraborti2019plan}, and more recently in multi-agent environments \cite{Kraus_2020}. 

In this work, we formulate contrastive explanations $\mathcal{E_{S}}$ for multi-robot task allocation, scheduling, and motion-planning solutions. We define this problem in relation to a team of heterogeneous robots, $R = \{r_1, r_2, ... r_n\}$, cooperating to execute a set of tasks, $T= \{t_1, t_2, ... t_m\}$, within a particular problem domain $\mathcal{D}$. Given $\mathcal{D}$, planning algorithm $\mathbb{A}$ solves the domain under constraints $\tau$ (time, utility, etc.), producing a solution $\mathcal{S}$ ($\mathbb{A}: \mathcal{D} \times \tau \rightarrow \mathcal{S}$). Consistent with prior work, we define a solution to the problem as $\mathcal{S} = \langle\mathcal{A}, \sigma, \mathcal{M}\rangle$, where $\mathcal{A: R \rightarrow}T$ is the allocation mapping robots to tasks, $\sigma$ is the task schedule, and $\mathcal{M}$ is the set of motion plans required for task execution {\cite{neville2021interleaved}}.  Our objective is to generate a natural language explanation that justifies the validity of $\mathcal{S}$ to the user, or else aids the user in correcting any errors that led to a suboptimal $\mathcal{S}$.

Given $\mathcal{S}$, we consider a human operator whose task is to evaluate the quality and correctness of the provided solution prior to execution.  In the presence of multiple robots with varying capabilities, as well as a diverse set of tasks, validating the solution requires the user to reason about i) the capabilities of each robot, ii) task requirements, and iii) scheduling constraints. We posit that a meaningful explanation of $\mathcal{S}$, $\mathcal{E_{S}}$, should consider each of these factors. 

In order to elicit an explanation from the system, we enable a human operator to specify a contrastive example, or \textit{foil}, that represents a foil solution, $\mathcal{S'}$.  In this work, we represent the foil as an alternate task allocation $\mathcal{A'}$, from which an alternative schedule, $\sigma'$, and motion plan, $\mathcal{M'}$, are automatically derived using $\mathbb{A}$. Our work directly generalizes to foils relating to alternate schedules or motion plans, as in {\cite{brandao2022taxonomy}}, as our explanations address each of these (Sec. {\ref{sec:method})}.

Given $\mathcal{S'}$, we define the contrastive explanation $\mathcal{E_{S}}$ under two scenarios:


\begin{enumerate}
    \item $\mathbb{A}: \mathcal{D} \times \tau \not\rightarrow \mathcal{S}'$ If $\mathcal{S'}$ is infeasible, $\mathcal{E_{S}}$ provides information about the cause. We define a solution $\mathcal{S}$ to be infeasible if the solution: i) assigns a robot to a task without meeting the task's minimum trait requirements, ii) violates a precedence constraint, or iii) has no valid motion plan solution.

    \item $\mathbb{A}: \mathcal{D} \times \tau \rightarrow \mathcal{S}'$, but $\mathcal{S} \equiv \mathcal{S}'$ or $\mathcal{S} > \mathcal{S}'$ In this case case, the user provides a feasible $\mathcal{S'}$ and $\mathcal{E_{S}}$ utilizes elements from $\mathcal{S}$ as reasoning for why
    $\mathcal{S} \equiv \mathcal{S}'$ or $\mathcal{S} > \mathcal{S}'$.
\end{enumerate}

\section{Multi-Robot Planning Problem Description}
\vspace{-.1cm}
\label{sec:prob_des}
To formulate a solution explanation $\mathcal{E_{S}}$, we require a formal definition of the multi-robot problem and its solution.  

\subsection{Problem Domain}
We consider a multi-robot scenario formulated by Neville et al. \cite{neville2021interleaved}, in which a multi-robot system provides a solution $S$ to a particular problem domain $\mathcal{D} = \langle \mathcal{Q}, \phi, \mathcal{T}, {Y^*} \rangle$, where
\begin{itemize}
    \item $\mathcal{Q}$ is the robot trait matrix describing the capabilities of each robot type.  A robot trait matrix $\mathcal{Q}$ is a column-wise concatenation of each robot's trait vectors. Specifically, considering a system with $N$ robots, a robot's traits can be defined as a continuous vector of length $U$,
\[q^{(i)} = [q_{1}^{(i)}, q_{2}^{(i)}, ... , q_{U}^{(i)}]\]
where $q_{u}^{i} \in R_{+}$ corresponds to the $u^{th}$ trait and the $i^{th}$ robot.
The $u^{th}$ trait represents a robot's characteristics, e.g., the robot has a robotic arm or a robot's carry capacity. Similar to Neville et al. \cite{neville2021interleaved}, we set $q_{u}^{i}$ to a positive real value or 0, depending on whether the $i^{th}$ robot does or does not possess the $u^{th}$ trait. In this manner, $Q$ is explicitly defined below.
\[\mathcal{Q} = [q^{(1)^T}, q^{(2)^T}, ... , q^{(N)^T}] \in \mathbb{R}^{N \times U}_{+}\]

   \item $\phi$ is a vector of length $N$ representing the speeds of $N$ robot types, 
\[\phi = [\phi_1, \phi_2, ... , \phi_N]\]
where $\phi_i \in \mathbb{R}_+$ corresponds to the $n^{th}$ robot's speed. 

   \item $\mathcal{T}$ is the task network, defined as a directed graph $\mathcal{G} = (\mathcal{E}, \mathcal{V})$, where each vertex in $\mathcal{V}$ is some task, $t_i \in T$. An edge in $\mathcal{E}$ connects two vertices in $\mathcal{V}$, shown by, $e=(t_i, t_j)$, where, $t_i, t_j \in \mathcal{V}$. This edge defines a precedence constraint, $t_i \prec t_j$, where task $t_i$ must be completed before task $t_j$ can start. 

   \item $Y^*$ is the desired trait matrix representing a column-wise concatenation of each task's trait requirements. Specifically, a task in $\mathcal{T}$ can be defined by a trait requirement vector of length $U$,
\[y^{(i)} = [y_{1}^{(i)}, y_{2}^{(i)}, ... , y_{U}^{(i)}]\]
where $y_{u}^{i} \in \mathbb{R}_{+}$ corresponds to the $u^{th}$ trait and the $i^{th}$ task. Similar to the trait vector, $q_{u}^{i}$, the value of $y_{u}^{i}$ is 0 if the $i^{th}$ task does not have that trait requirement, and otherwise is a positive real value. In this manner, $Y^*$ is explicitly defined as
\[Y^* = [y^{(1)^T}, y^{(2)^T}, ... , y^{(M)^T}] \in \mathbb{R}^{M \times U}_{+}\]

\end{itemize}

\subsection{Solution Specification}
Given the above problem formulation $\mathcal{D}$, a multi-robot planner finds solution $\mathcal{S} = \langle\mathcal{A}, \sigma, \mathcal{M}\rangle$ in which $\mathcal{A}$, $\sigma$, and $\mathcal{M}$ represent a task allocation matrix, schedule and motion plan, respectively. Specifically, the task allocation matrix $\mathcal{A}$ is a $M \times N$ matrix, where $M$ is the number of tasks and $N$ the number of robots. If an element $A^{n}_{m}$ = 1, then robot $n \in \{1, ..., N\}$ is allocated to task $m \in \{1, ..., M\}$. Otherwise, if $A^{n}_{m}$ = 0, then robot $n$ is not allocated to task $m$. The schedule $\sigma$ assigns a start and end time for each task in the task network $\mathcal{T}$. Note, $\sigma$ ensures that precedence constraints defined in $\mathcal{T}$ and  mutex constraints in $\mathcal{M}$ are met. Finally, the motion plan $\mathcal{M}$ represents as a finite set of motion plans for each robot with their assigned tasks defined in $\mathcal{A}$. 


The search space for $\mathcal{S}$ is large, spanning hundreds of potential allocations \cite{messing2022grstaps}.  Our objective is to provide explanations to the user that reveal why $\mathcal{S}$ is a valid solution. Alternately, if errors exist in the domain specification $\mathcal{D}$ (e.g., incorrect robot speed parameters, errors in the task network specification, etc), it is helpful if the explanations can help the operator to identify those errors.  In the following section we discuss our approach for generating such explanations.

\section{CE-MRS Framework}
\vspace{-.1cm}
\label{sec:method}

\begin{figure} [!t]
\centering
\includegraphics[width=8cm]{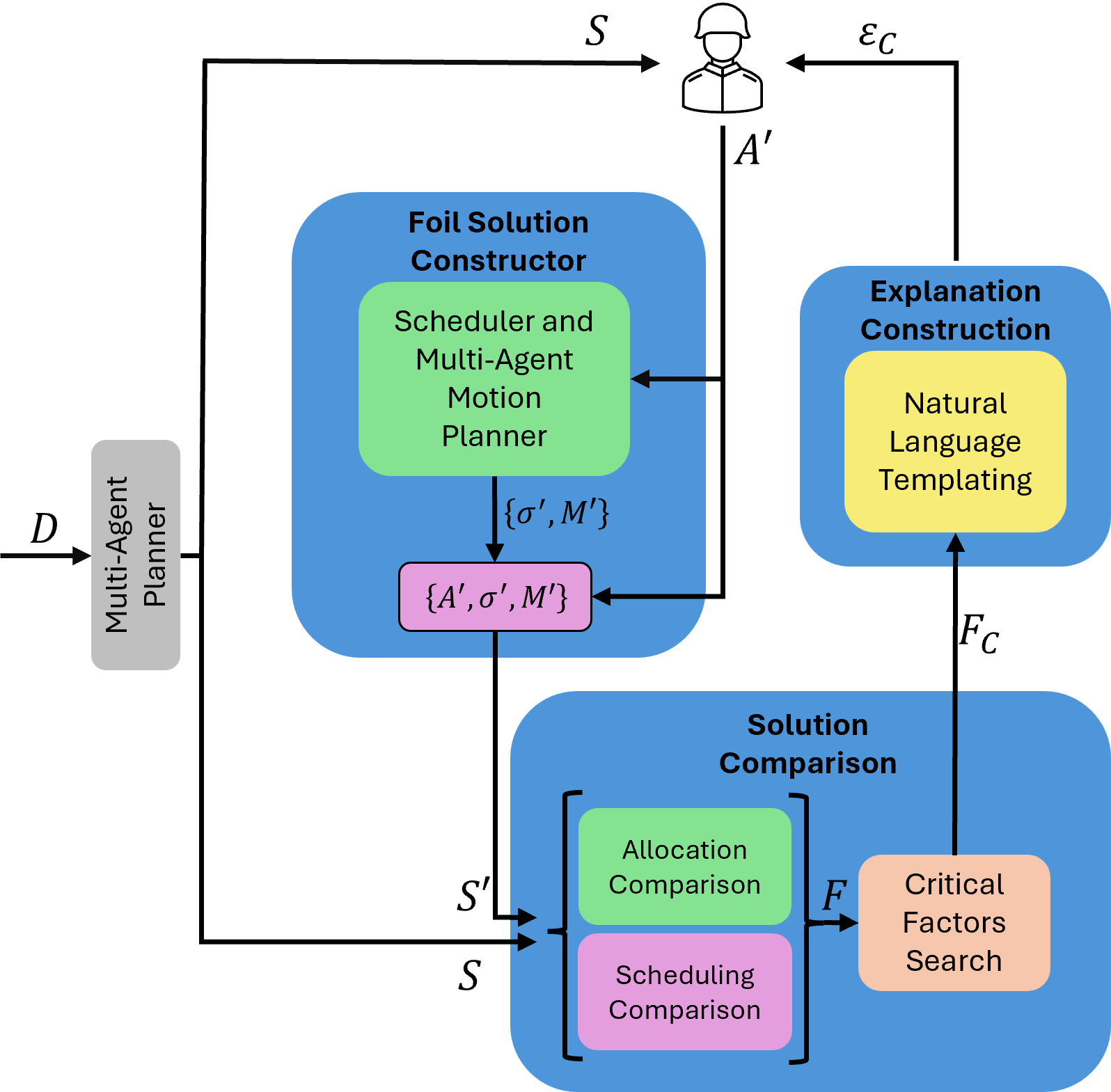}
\caption{CE-MRS Framework Diagram}
\label{fig:sys_overview}
\vspace{-1em}
\end{figure}

Figure \ref{fig:sys_overview} presents an overview of the CE-MRS framework that provides contrastive explanations, $\mathcal{E_{S}}$, for a multi-robot system's solution $\mathcal{S}$, given a problem domain $\mathcal{D}$.  The explanation process begins with a human operator, who receives $\mathcal{S}$ for a given $\mathcal{D}$.  If the solution is unexpected or unclear, the user is able to gain more information about $\mathcal{S}$ by posing a foil in the form of $\mathcal{A}'$, an alternative task allocation. The CE-MRS framework then provides $\mathcal{E_{S}}$ which compares the user's proposed solution $\mathcal{S}'$ (based on $\mathcal{A}'$) with $\mathcal{S}$. Specifically, CE-MRS consists of three modules to provide $\mathcal{E_{S}}$.
The \textit{Foil Solution Constructor} module takes the user's foils, in the form of $\mathcal{A'}$, and generates a corresponding $\mathcal{S'} = \langle \mathcal{A'}, \sigma', \mathcal{M'}\rangle$.
The \textit{Solution Comparison} module then finds $F$, the set of all differences between $\mathcal{S'}$ and $\mathcal{S}$, and further filters $F$ into a subset $F_{C}$, containing only the top critical differences between $\mathcal{S'}$ and $\mathcal{S}$. Finally, the \textit{Explanation Constructor} module templates $F_{C}$ into a formatted natural language explanation $\mathcal{E_{S}}$. Below we further detail each module within CE-MRS.

\subsection{Constructing the Foil Solution}
For a given solution $\mathcal{S}$, the user can ask foils about the system's task allocation $\mathcal{A}$. Specifically, users can ask questions in the form, \textit{``Why is $r_{i}$ not assigned to $t_{j}$?"} where $r_{i}$ represents a given robot and $t_{j}$ represents a particular task. The user can pose one or more of these questions, and the set of questions constitute $\mathcal{A'}$. To generate a counterfactual solution $\mathcal{S}'$, CE-MRS leverages $\mathcal{A'}$ to automatically construct foil schedule $\sigma'$ and foil motion plan $\mathcal{M}'$. Specifically, $\mathcal{A}'$ serves as input to a multi-robot planner's scheduler to generate  $\sigma'$, and $\sigma'$ serves as input to a multi-robot planner's motion planner to generate $\mathcal{M}'$. 

In this work, we utilize the ITAGS algorithm \cite{neville2021interleaved} to generate $\mathcal{S}$ and $\mathcal{S'}$, although CE-MRS can be adapted to work with other representations and multi-agent planning techniques that model task dependencies, allocations, schedules, and durations.  
Additionally, while in this work we restrict the user's foil to $\mathcal{A'}$, the CE-MRS framework easily generalizes to scheduling foils ($\sigma'$) and motion planning foils ($\mathcal{M'}$).

\subsection{Solution Comparison}
Given $\mathcal{S}$ and $\mathcal{S}'$, the solution comparison module in CE-MRS computes the set of factors $F$ through which $\mathcal{S}$ and $\mathcal{S}'$ differ. Specifically, CE-MRS computes differences in task allocations ($\mathcal{A}$ vs $\mathcal{A'}$) and schedules ($\sigma$ vs $\sigma'$). For complex multi-robot systems, $|F|$ may be large, and it is important to ensure that $\mathcal{E_{S}}$ remains interpretable to the user. Prior work has demonstrated that abstracting explanations to include only the most important factors leads to improved user understanding \cite{lai2023selective,hudon2021explainable}. Given these findings, the solution comparison module additionally employs a threshold to find the subset $F_{C} \subseteq F$ that represent in the most important factors through which $\mathcal{S}$ and $\mathcal{S}'$ differ. Below we detail the allocation comparison, scheduling and motion planning comparison, and the critical factors filtering algorithm.

\subsubsection{\textbf{Allocation Comparison}} 
Recall that $\mathcal{A}$ represents a $M \times N$ task allocation matrix where $M$ represents the number of tasks and $N$ represents the number of robots. The allocation comparison algorithm performs a column-wise comparison between $\mathcal{A}_{m}$ and $\mathcal{A'}_{m}$ to find $F_{\mathcal{A}}$, the allocation-related factors in which $\mathcal{A}_{m}$ and $\mathcal{A'}_{m}$ differ. Specifically, when $\mathcal{A}_{m} \neq \mathcal{A}_{m}'$, 
$F_{\mathcal{A}} = (m, n, n') \forall m \in \{1, ..., M\}$, where $m$ is the allocated task, $n$ is the allocated robot in $\mathcal{A}$, and $n'$ is the allocated robot in $\mathcal{A}'$.

\subsubsection{\textbf{Scheduling and Motion Planning Comparison}}
The scheduling comparison algorithm computes the differences between $\sigma$ and $\sigma '$ based on several characteristics important to scheduling. Specifically, the algorithm considers the percent differences between $\sigma$ and $\sigma '$ in terms of overall makespan $\lambda$, task time $\beta_{m}$ where $m \in \{1, ..., M\}$, and robot makespan $\alpha_{n}$ where $n \in  \{1, ..., N\}$. Note, overall makespan refers to the total time taken for all tasks to be completed, task time refers to the time taken for a certain task to be completed, and robot makespan refers to the time taken for a given robot to complete its assigned tasks. For all $M$ tasks and $N$ robots, the schedule-related factors, $F_{\sigma} = \{\lambda, \beta, \alpha \} $, where $\beta = \{\beta_{0}..\beta_{M}\}$ and $\alpha = \{\alpha_{0}..\alpha_{N}\}$.

\begin{table*}[t]
\vspace{0.5em}
\resizebox{\textwidth}{!}{%
\begin{tabular}{|c|l|l|}
\hline
Error Type & \multicolumn{1}{c|}{CMAoE} & \multicolumn{1}{c|}{CE-MRS}                                                               \\ \hline
Q Error    & Error in foil. Cannot Run & • \textbf{ambulance({[}2500, 1, 1{]})} and dumptruck({[}5000, 0, 1{]}) can work D1({[}600, 0, 1{]}) \\ \hline
Y* Error   & Error in foil. Cannot Run & • ambulance({[}2500, 1, 0{]}) and dumptruck({[}5000, 0, 1{]}) can work \textbf{D1({[}600, 1, 1{]})} \\ \hline
$\phi$ Error &
  \begin{tabular}[c]{@{}l@{}}\\User's solution is 16.71 min. longer described by the table:\\  \\  Robot           \hspace{1cm} dumptruck1   \hspace{1cm}   ambulance1\\ Decrease    \hspace{0.5cm}  D3,(11.07 min.)      \hspace{1.2cm}         -\\ Increase \hspace{0.4cm}  \textbf{D1, D2,(24.55 min.)} \hspace{0.2cm} H1,(1.68 min.)\end{tabular} &
  \begin{tabular}[c]{@{}l@{}}• ambulance({[}2500, 1, 0{]}) and dumptruck({[}5000, 0, 1{]}) can work D1({[}600, 0, 0{]})\\ User's solution takes 32\% more time: 45.55 minutes$\rightarrow$63.4 minutes\\     • dumptruck1 takes 32\% more time\\    \hspace{0.5cm}     • D1 takes 154\% more time: \textbf{ambulance(40.0m/s)}$\rightarrow$dumptruck(4.0m/s)\\    \hspace{0.5cm}     • D3 takes 25\% less time\\     • ambulance1 takes 29\% less time\\    \hspace{0.5cm}     • H1 takes 45\% more time\end{tabular} \\ \hline
\multicolumn{1}{|l|}{Q, Y*, and $\phi$ Error} &
  \begin{tabular}[c]{@{}l@{}}\\User's solution is 16.16 min. longer described by the table:\\ \\   Robot           \hspace{1cm} dumptruck1    \hspace{1cm}  ambulance1\\ Decrease   \hspace{0.5cm}   D3,(11.22 min.)      \hspace{1.2cm}         -\\ Increase \hspace{0.4cm}  \textbf{D1, D2,(24.15 min.)} \hspace{0.2cm} H1,(1.72 min.)\end{tabular} &
  \begin{tabular}[c]{@{}l@{}}Task and Robot Capabilities Comparison:\\     • ambulance({[}2500, 1, 0{]}) and \textbf{dumptruck({[}5000, 1, 1{]})} can work \textbf{D1({[}600, 0, 0{]})}\\ User's solution takes 32\% more time: 45.65 minutes$\rightarrow$63.22 minutes\\     • dumptruck1 takes 32\% more time\\     \hspace{0.5cm}    • D1 takes 153\% more time: \textbf{ambulance(40.0m/s)}$\rightarrow$dumptruck(4.0m/s)\\    \hspace{0.5cm}     • D3 takes 26\% less time\\     • ambulance1 takes 30\% less time\\     \hspace{0.5cm}    • H1 takes 44\% more time\end{tabular} \\ \hline
\end{tabular}%
}
\caption{Example explanations generated for errors in $\mathcal{Q}$, $\phi$, and $Y^*$ for both study conditions. 
The bold text shows the errors which the explanations are highlighting for the user.}
\label{tab:explanation-example}
\vspace{-1.5em}
\end{table*}

\subsubsection{\textbf{Critical Factors Filtering}}
From the set of factors $F$, we perform empirically-driven thresholding to extract the top contributing factors $F^{C} \subseteq F$. Recall that $F = \{F_{\mathcal{A}}$, $F_{\sigma}\}$, denoting factors related to differences between task allocation and scheduling, respectively. In this manner, $F^{C} = \{F^{C}_{\mathcal{A}}, F^C_{\sigma}\}$, where $F^{C}_{\mathcal{A}} \subseteq F_{\mathcal{A}}$ and $F^{C}_{\sigma} \subseteq F_{\sigma}$. 

To find $F^C_{\sigma}$, we define a threshold, $Z$, such that any value of a factor $f_{\sigma} \in F_{\sigma}$ above $Z$ is determined to be a critical factor. In our work, $Z$ is computed through an ablation study in which we simulate multiple user foils with a varying value of $Z$, and find the
point at which the rate of change decreases significantly (in our case, $Z=0.1$). Similarly, in this work, we set 
$F^{C}_{a} = F_{a}$ and consider all allocation-related factors as critical factors. 

\subsection{Explanation Construction}
Given the top contributing factors $F^{C}$, the explanation construction module produces $\mathcal{E_{S}}$ by templating $F^{C}_{\mathcal{A}}$ and $F^{C}_{\sigma}$ into a natural language format. Specifically, we translate $F^{C}_{\mathcal{A}}$ into natural language by constructing a sentence that states a given task $m$ is capable of being worked by robots $n$ and $n'$, in which the task requirement $Y^{*}_{m}$ and robot traits $\mathcal{Q}_n$ and $\mathcal{Q}_{n'}$ are appended. Similarly, $F^{C}_{\sigma}$ is translated into natural language by enumerating the percent differences in $\{\lambda, \beta, \alpha \}$, in which it reveals $\phi_{n}$ as a possible reason for why $\beta_{m}$ is significantly different between robots $n$ and $n'$. Examples of $\mathcal{E_{S}}$, given $F_{C} = \{F_{\mathcal{A}}$, $F_{\sigma} \}$ is provided in Table \ref{tab:explanation-example}. 

Note, while prior work has considered providing explanations for task allocation, they only consider either why $\mathcal{S} > \mathcal{S'}$ or why $\mathcal{S}$ is valid. Specifically in Zehtabi et. al \cite{zehtabi2023contrastive}, the authors consider justifying one solution over another by an objective function (i.e. utility, resources, time). In Zahedi et. al \cite{zahedi2023why}, the authors use a negotiation-aware approach of explaining why one solution was chosen, by arguing why some agent would reject some foil solution. In our work, we address both feasibility and optimally for some objective variable in $\mathcal{E_S}$, and we utilize information from the input ($\mathcal{D}$) and output ($\mathcal{S}$ and $\mathcal{S'}$) to formulate our explanations. 


\section{Experimental Design}
\vspace{-.1cm}

Our study objectives are to answer two research questions:
\begin{itemize}
    \item \textit{RQ1: Can human operators detect when a multi-robot planning solution is incorrect?}  This question is critical because users are only likely to ask ``Why did/didn't the robot...?'', and engage with the explanation system, if they can independently assess plan solution quality and detect unexpected results.  Thus, we first assess participant ability to correctly assess plan solution quality.
    
    \item \textit{RQ2: How well do contrastive explanations enable human operators to identify and correct any errors within the multi-robot problem domain specification?} We particularly examine the scenario in which some other (hypothetical) human team member encodes the domain representation $\mathcal{D}$, consisting of its components $\mathcal{Q}$ (robot trait matrix), $\phi$ (robot speed vector), $\mathcal{T}$ (task network), and $Y^*$ (desired trait matrix).  In our study, we corrupt $\mathcal{D}$ in some subset of study scenarios, and then measure the performance of our study participants in identifying and correcting the introduced errors.  The participant objective is to make $\mathcal{D}$ equal to $\mathcal{D^*}$.
\end{itemize}

To examine the above research questions, we conducted a two-way, between subjects study in which participants analyzed multiple simulated, emergency-response scenarios inspired by \cite{kitano1999robocup} and \cite{zhao2016multitaskalloc}.  For each scenario, the participant was first asked whether the proposed plan solution was sound (RQ1).  If the participant believed the solution was sound, they were allowed to move on to the next scenario.  Alternately, if the participant felt that errors were likely, they were allowed to engage with the explanation system and address any perceived errors in $\mathcal{D}$ (RQ2).  Once complete, participants could either state that they believed the final solution was correct ($\mathcal{D} == \mathcal{D^*}$), or that errors remained but they did not know how to fix them.  Participants were randomly split into two conditions based on the type of explanation received:
\begin{enumerate}
    \item \textbf{CE-MRS (ours)}: participants received explanations from the CE-MRS Framework, detailing whether $\mathcal{S'}$ is feasible, and summarizing the top contributing factors through which $\mathcal{S}$ and $\mathcal{S}'$ differ.
    \item \textbf{CMAoE (baseline)}: received contrastive tabular explanations, as in \cite{zehtabi2023contrastive}, in which a user asks ``why does $S$ not enforce property $P$'', and CMAoE generates $S'$ where $P$ is enforced, while minimizing the difference between $S$ and $S'$. 
    
\end{enumerate}
Note, we chose CMAoE \cite{zehtabi2023contrastive} as the closest related baseline to our work as this approach similarly supports task-allocation foils $\mathcal{A'}$, from which schedules and motion plans can be derived. We do not consider scheduling-only explanation methods \cite{ludwig2018scheduling,cyras2019scheduling} since these do not incorporate task allocation into their frameworks. 
Additionally, as we use ITAGS \cite{neville2021interleaved} for our underlying MAP, we do not consider AITA \cite{zahedi2023why} as a comparable baseline because AITA relies on their task allocation framework to generate their explanations, which makes an unfair comparison with the other evaluated methods using ITAGS. 
Table \ref{tab:explanation-example} provides example explanations from CE-MRS and CMAoE for different combination of errors in $\mathcal{Q}$, $Y^*$, and $\phi$. 

\begin{figure*}[t!]
    \vspace{0.1em}
    \centering
    \includegraphics[width=0.65\textwidth]{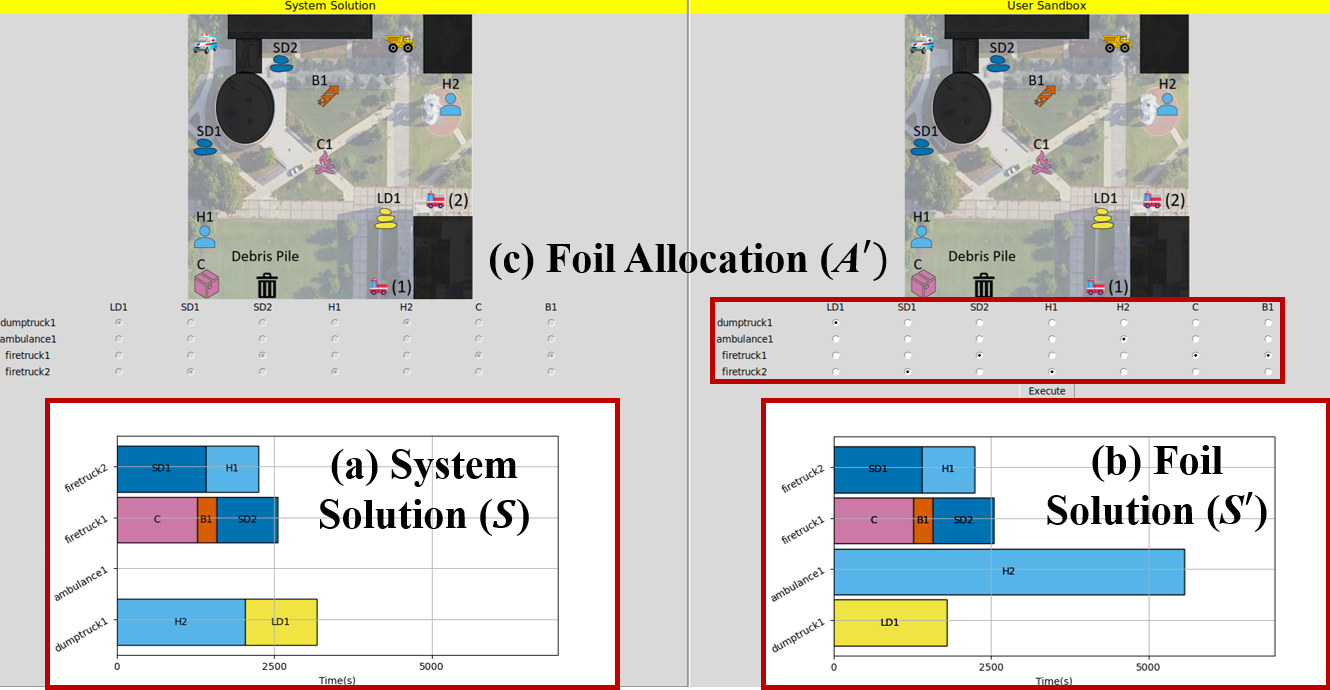}
    \caption{Our user interface, in which the System Solution Panel (left) visualizes the system solution $\mathcal{S}$ for domain definition $\mathcal{D}$ (a). The User Sandbox Panel (right) visualizes a user's foil solution $\mathcal{S'}$ (b) for their foil allocation $\mathcal{A'}$ (c).}
    \label{fig:system_panel}
    \vspace{-1em}
\end{figure*}

\subsection{Study Design}
Each study participant was asked to complete 6 independent experimental scenarios.  In each scenario, the participant was provided with $r$ robots and was asked to generate a solution $\mathcal{S}$, including an allocation, schedule and motion plan, for $k$ tasks. We maintained a consistent number of robots $r=4$ and tasks $k=7$ across the scenarios
\footnote{A pilot study that explored varying number of robots and tasks showed that $r=4$ and $k=7$ required significant user effort. We kept these numbers fixed since changing the difficulty of the problem across scenarios conflates results with varying the types of errors the users were exposed to.}
, while varying the types/number of errors injected in the problem definition. Robots used in the study were 1 dumptruck, 2 firetrucks, and 1 ambulance, each with traits described in Table {\ref{tab:robot-traits}}. Tasks included in the study were 1 large debris, 2 small debris, 2 rescue humans, 1 setup camp, and 1 defuse bomb. The initial robot and task configurations differed in each scenario (see example map in Figure {\ref{fig:system_panel}}), leading to different solutions.



\begin{table}[]
\begin{tabular}{|c|c|c|c|}
\hline
 Traits                                                            & Dumptruck & Firetruck & Ambulance \\ \hline
\begin{tabular}[c]{@{}c@{}}Carrying\\ Capacity (lbs.)\end{tabular} & 5000      & 1500      & 2500      \\ \hline
Stretcher                                                      & -         & -         & \checkmark         \\ \hline
\begin{tabular}[c]{@{}c@{}}Robotic\\ Arm\end{tabular}          & -         & \checkmark   & \checkmark         \\ \hline
Forklift                                                       & \checkmark         & \checkmark         & -         \\ \hline
Speed (m/s)                                                        & 4         & 7         & 8         \\ \hline
\end{tabular}
\centering
\caption{List of robots and their traits, including their speeds. All of these traits can be corrupted with an error.}
\label{tab:robot-traits}
\vspace{-1em}
\end{table}

\begin{table}[]
\begin{tabular}{|c|c|c|c|c|c|}
\hline
Requirements
& \begin{tabular}[c]{@{}c@{}}Large \\ Debris\end{tabular} 
& \begin{tabular}[c]{@{}c@{}}Small\\ Debris\end{tabular}
& \begin{tabular}[c]{@{}c@{}}Rescue\\ Human\end{tabular} 
& \begin{tabular}[c]{@{}c@{}}Setup\\ Camp\end{tabular} 
& \begin{tabular}[c]{@{}c@{}}Defuse\\ Bomb\end{tabular} \\ \hline
\begin{tabular}[c]{@{}c@{}}Carrying\\ Capacity (lbs.)\end{tabular} 
& 4200 & 500-1200 & 100-200 & 2000 & - \\ \hline
\begin{tabular}[c]{@{}c@{}}Stretcher\end{tabular}                   & - & - & \checkmark & - & - \\ \hline
\begin{tabular}[c]{@{}c@{}}Robotic Arm\end{tabular}                & - & \checkmark & - & \checkmark & \checkmark \\ \hline
\begin{tabular}[c]{@{}c@{}}Forklift\end{tabular}                    & \checkmark & \checkmark & - & - & - \\ \hline
Preconditions                                                                 & - & - & \begin{tabular}[c]{@{}c@{}}Setup\\ Camp\end{tabular}  & - & - \\ \hline
\end{tabular}
\centering
\caption{List of task trait requirements and preconditions. All of the tasks' requirements can be corrupted with an error.}
\label{tab:tasks}
\vspace{-1em}
\end{table}


For each scenario, participants were first presented with a plan solution $\mathcal{S}$ and were asked whether they thought it was correct (RQ1).  If the participant suspected errors were present, they then leveraged the explainability interface to generate counterfactual explanations, and were able to make corrections to the underlying problem specification $\mathcal{D}$ (RQ2).  The study was broken up into the two stages: \footnote{Scenarios are composed of a tuple of errors, (Robot Errors, Task Errors, Speed Errors). The six scenarios' errors are: (0,0,0), (3,1,1), (0,0,0), (2,2,1), (0,5,0), (3,2,0)}


\smallskip
\noindent{\underline{\textit{Familiarization}}}. Users first received a tutorial in the domain specification, including definition of each of the components of $\mathcal{D}$ (as defined in Sec. \ref{sec:prob_des}). They then received a walk-through of how to interpret the explanations from their assigned study conditions, completed a 5-question assessment to validate their understanding, and then completed two practice scenarios to ensure familiarity with the interface. 

\smallskip
\noindent{\underline{\textit{Assessment}}}.
Participants were presented with six scenarios; Figure \ref{fig:system_panel} provides a visualization of the study interface.  Participants were first shown the system's solution $\mathcal{S}$ (Figure \ref{fig:system_panel}a) in the System Solution Panel, and were asked if the solution appeared correct (RQ1).  All participants received the same six scenarios in random order to account for learning effects. In two scenarios, $\mathcal{D} = \mathcal{D}^*$, and there were no errors. The remaining scenarios had five errors each, either distributed among only robot traits and speeds, task requirements or both.

If participants believed $\mathcal{S}$ was incorrect due to errors in $\mathcal{D}$, they were provided with the User Sandbox Panel, which enabled them to generate $\mathcal{S'}$ (Figure \ref{fig:system_panel}b) and the associated explanation, by asking $\mathcal{A'}$ (Figure \ref{fig:system_panel}c). Users could then update $\mathcal{D}$ to resolve any identified errors. This process would be repeated until the participant believed $\mathcal{D} = \mathcal{D}^*$ (the solution was correct), or until they gave up by indicating that they wanted to change scenarios but that they believed unknown errors remained.  To discourage participants from reaching a trivial solution by manually re-entering all the matrix values in $\mathcal{D}$, we limited the interface to require each value to be changed individually (e.g., set $\phi(r1)=10$, the velocity of robot $r1$, to $10$).




\subsection{Metrics}
We utilised the following metrics to evaluate our study:
\begin{enumerate}
    \item \textbf{{User Scenario Classification Prior to Explanations (\textit{IDP})}}: Users' accuracy score for identifying if $\mathcal{D}$ is valid prior to viewing explanations.
    

    
    \item \textbf{\textbf{Remaining Robot Trait Errors 
 (\textit{RTE\%})}}: Percentage of errors in the robot trait matrix ($\mathcal{Q}$) that remain unresolved at the end of the scenario.
    
    \item \textbf{\textbf{Remaining Task Requirement Errors (\textit{TRE\%})}}: Percentage of errors within the desired trait matrix ($Y^*$) that remain unresolved at the end of the scenario.
    
    \item \textbf{\textbf{Remaining Robot Speed Error (\textit{RSE\%})}}: Percentage of errors within the robot speeds ($\phi$) that remain unresolved at the end of the scenario.

    \item \textbf{User Efficiency (\textit{UE})}: Ratio between the number of repair actions and the total remaining errors, where repair attempts refers to changes made to $\mathcal{D}$; this metric is an approximation for user effort.

    \item \textbf{{User Scenario Classification After Explanations (\textit{IDA})}}: Users' accuracy score for identifying if $\mathcal{D}$ is valid after viewing explanations. 


\end{enumerate}

\subsection{Participants}
We recruited 22 participants from a US university. Participants were randomly placed into a study condition, resulting in 11 participants in each study condition. Our participants included 17 males, 9 females, and 1 non-disclosed, all of whom are over the age of 18 (M=24.7, SD=2.29). The study took 1hr - 1.5hrs, and participants were compensated \$15.

\section{Results}
\vspace{-.1cm}

Given that participants' \textit{RTE\%}, \textit{TRE\%}, and \textit{RSE\%} do not follow a normal distribution (Shapiro-Wilk's Test, p $<$ 0.05), we analyze statistical significance between study conditions using the non-parametric Wilcoxon rank-sum test. We also provide a qualitative analysis of participants' \textit{UE} scores.

\subsection{Scenario Classification Prior to Explanation}
We first evaluate the \textbf{IDP} metric to determine whether users can correctly identify when errors exist in the system's original solution $\mathcal{S}$.  Given that $\mathcal{S}$ is presented with no accompanying explanations, we expect little difference in \textbf{IDP} scores across conditions.  Our analysis confirms this hypothesis, with IDP values of 69.7\% and 77.3\% for CMAoE and CE-MRS, respectively, showing that operators had similar ability to assess solution accuracy across study conditions.

\subsection{Operator Ability to Resolve Errors}
\begin{figure}[h]
\centering
\begin{subfigure}[b]{0.15\textwidth}
  \includegraphics[width=\textwidth]{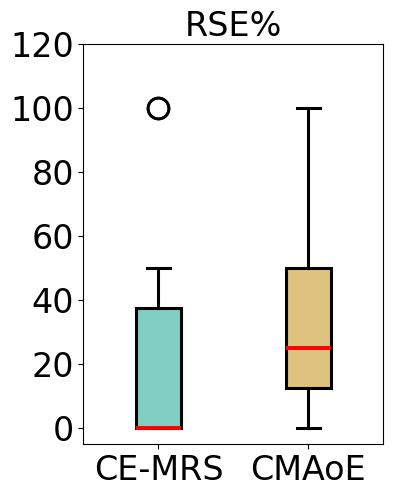}
    \caption[]%
     {}
\end{subfigure}
\begin{subfigure}[b]{0.15\textwidth}
  \includegraphics[width=\textwidth]{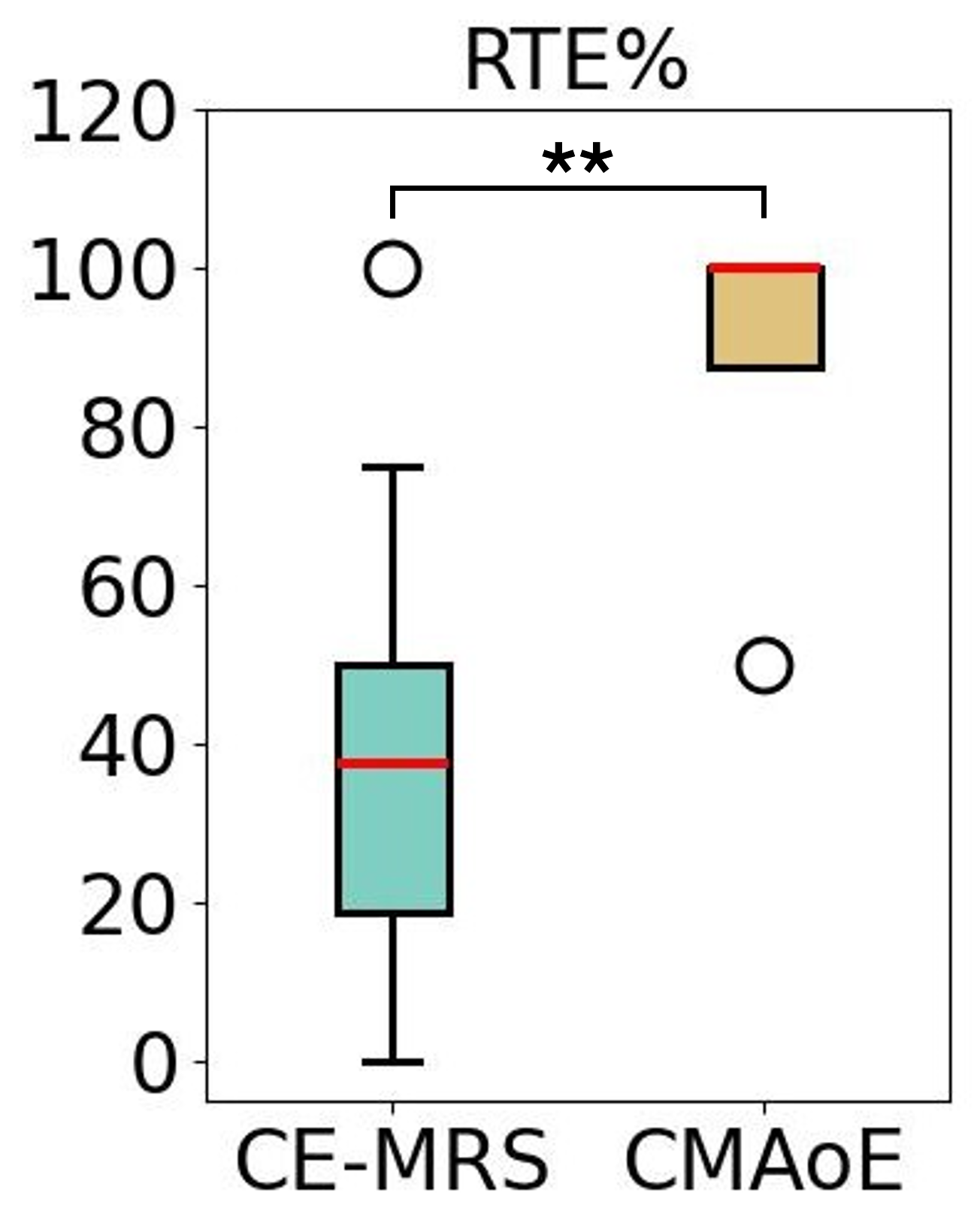}
  \caption[]%
  {}
\end{subfigure}\quad
\begin{subfigure}[b]{0.15\textwidth}
  \includegraphics[width=\textwidth]{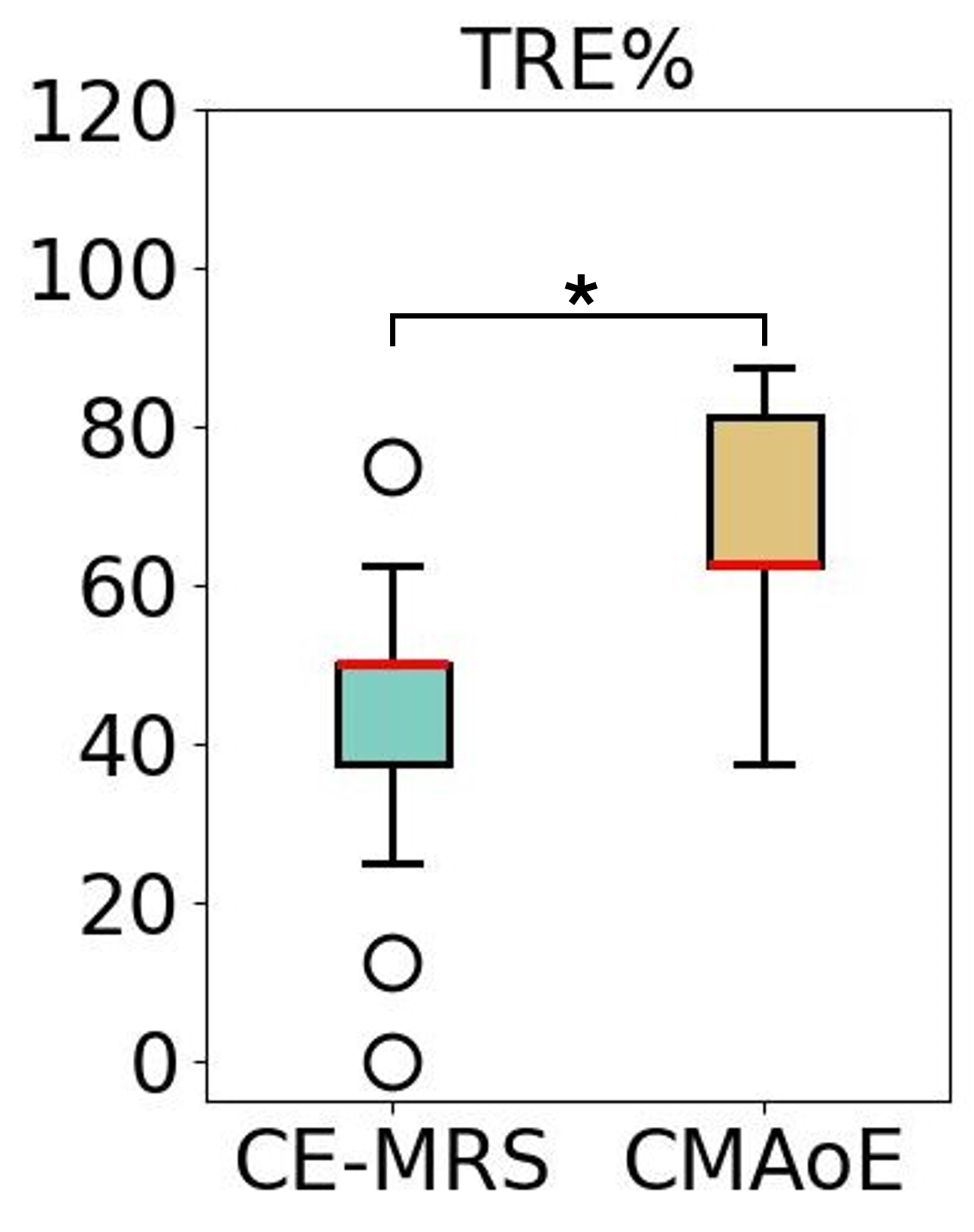}
    \caption[]%
     {}
\end{subfigure}
\caption{RSE\%, RTE\%, and TRE\% metrics per study condition, in which a lower value is better. Statistical significance
is reported as: * p$<$0.01, ** p$<$0.001}
\label{fig:remainingerror}
\vspace{-0.5em}
\end{figure}

Next, we examine user effectiveness in resolving errors in the domain definition by examining the total number, and type, of errors that remained once users submitted their final solution.  Figure \ref{fig:remainingerror} presents remaining error percentage in $\mathcal{D}$ for scenarios in which the domain description did not match the ground truth. We examine three types of unresolved errors, \textbf{RSE\%}, \textbf{RTE\%}, \textbf{TRE\%}, which relate to errors in $\phi$, $\mathcal{Q}$, and $Y^*$, respectively. Lower values correspond to better performance, as the user has corrected more errors in $\mathcal{D}$.

In Figure \ref{fig:remainingerror}a, we observe that participants identify and correct errors in $\phi$ at similar rates in CE-MRS and CMAoE, and we find no statistical difference between the conditions. 
However, we observe that CE-MRS has a lower variance of \textbf{RSE\%} than the baseline, indicating greater consistency among operators.  This is likely due to the fact that CE-MRS explicitly reveals elements of $\phi$ from $\mathcal{D}$ in $\mathcal{E_S}$.

For the \textbf{RTE\%} and \textbf{TRE\%} metrics (Figure \ref{fig:remainingerror}b and Figure \ref{fig:remainingerror}c), we observe that CE-MRS significantly outperforms CMAoE in assisting users to identify and correct $\mathcal{Q}$ errors, \textbf{RTE\%} ($p<0.001$), and $Y^*$ errors, \textbf{TRE\%} ($p<0.01$). This is because CE-MRS selectively reveals information from the task allocation, scheduling, and motion planning sub-modules to explain the system's reasoning for a particular solutions, resulting in a higher user understanding of the system's solution. In the first two rows of Table \ref{tab:explanation-example}, we show CE-MRS explanation examples and highlight in bold where we explicitly reveal information from $\mathcal{Q}$ and $Y^*$.

\subsection{User Efficiency}
\begin{figure}[h]
    \centering
    \includegraphics[width=8cm]{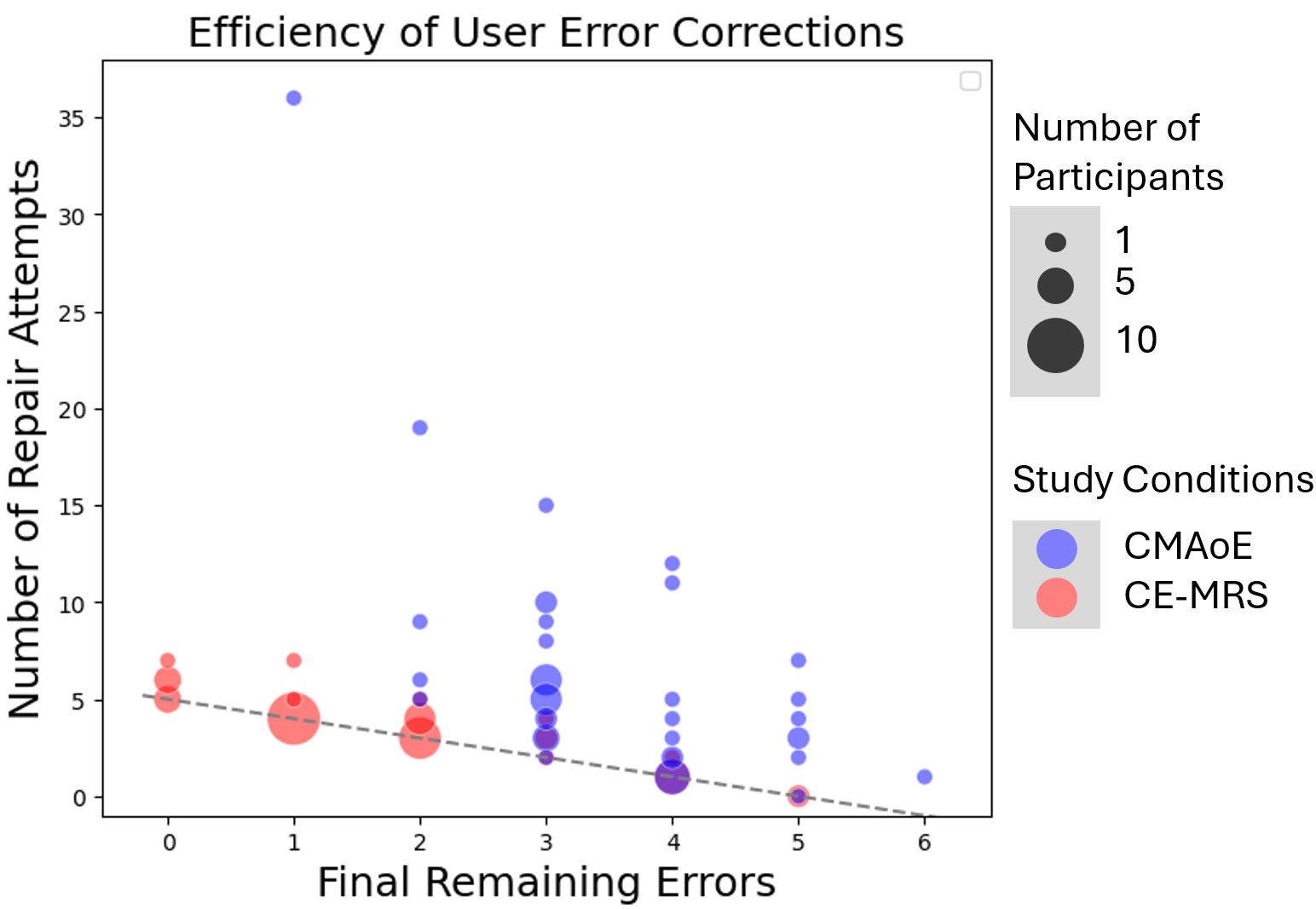}
    \caption{Efficiency of users' error corrections of $\mathcal{D}$. Points further left are scenarios with more errors corrected; points closer to the line represent more efficient corrections.}
    \label{fig:efficiency}
    \vspace{-0.5em}
\end{figure}

In this section, we examine how efficient users were in resolving errors.
Figure \ref{fig:efficiency} visualizes the \textbf{UE} metric separated by study condition for how efficient users are in correcting errors by making changes to $\mathcal{D}$. In this graph, the dashed line represents a perfectly efficient user that makes no redundant changes to $\mathcal{D}$. Data points above the line correspond to scenarios in which users make more repair attempts than errors corrected. Scenarios with more corrected errors have the least Final Remaining Errors (lower x-axis values).  

In Figure \ref{fig:efficiency}, we observe that users with CE-MRS explanations follow the dashed line more closely than participants given CMAoE explanations, indicating that our CE-MRS explanations help users correct
errors in $\mathcal{D}$ more efficiently.
Additionally, users in the CE-MRS condition have fewer Final Remaining Errors. In fact, no participants in the CMAoE condition ended a scenario without remaining errors. This highlights the importance of selectively revealing information from the system input, $\mathcal{D}$, when comparing $\mathcal{S}$ and $\mathcal{S}'$ or explaining a solution's feasibility.

\subsection{Perceived Correctness of Final Solution}
As shown in the previous section, 75\% of CE-MRS and 100\% of CMAoE scenarios ended with at least one error remaining in $\mathcal{D}$.  In this section, we examine whether participants were aware that errors remained, or whether they believed that the solution was fully correct, as measured by the \textbf{IDA} metric.  Our results show that, similar to \textbf{IDP}, participants showed similar ability to evaluate solution accuracy across conditions, with \textbf{IDA} values of 22.7\% and 18.2\% for CMAoE and CE-MRS, respectively.  We conclude that while CE-MRS explanations allow users to resolve statistically significantly more errors, neither type of explanation improved user ability to assess correctness of an overall multi-robot plan solution. This outcome is not entirely surprising, as neither explanation approach is designed to evaluate optimality of $\mathcal{S}$, so other techniques should be designed to address this challenge.

\subsection{Conclusions}
In summary, our results demonstrate that CE-MRS significantly improves a human operator's ability to identify and resolve errors in the problem formulation relating to robot capabilities (RTE\%) and task requirements (TRE\%), while performing on-par with prior methods on resolving errors in robot capabilities (RSE\%). 
 In total, participants in the CE-MRS condition were able to resolve $62.7\pm23.5$\% of errors, compared to only $29.5\pm9.40$\% when using CMAoE explanations.  Participants exposed to CE-MRS explanations were also more efficient in resolving errors, with CE-MRS participants making only $1.33\pm0.60$ extraneous corrections, on average, compared to $5.24\pm5.63$ extraneous corrections in CMAoE.  Participants in both conditions showed similar ability to assess overall solution correctness.   

\section{Limitations \& Future Work}
\label{sec:conclusions}




While our results demonstrate strong advances in improving the interpretability of multi-robot systems, there there are numerous opportunities for future work.
First, we determined the threshold $Z$ through an ablation study for our particular domain; future work should explore more generalizable methods for filtering the factors $F$ to the critical factors $F^C$.
Additionally, future work should investigate how to improve operator ability to correctly assess the correctness of a solution. Finally, more longitudinal field studies are needed to assess explanability techniques for multi-robot systems.

\bibliographystyle{IEEEtranS}
\bibliography{references}

\end{document}